# Stability of meanings versus rate of replacement of words: an experimental test

Michele Pasquini & Maurizio Serva
Dipartimento di Ingegneria e Scienze dell'Informazione e Matematica, Università dell'Aquila, L'Aquila, Italy.

The words of a language are randomly replaced in time by new ones, but it has long been known that words corresponding to some items (meanings) are less frequently replaced than others. Usually, the rate of replacement for a given item is not directly observable, but it is inferred by the estimated stability which, on the contrary, is observable. This idea goes back a long way in the lexicostatistical literature, nevertheless nothing ensures that it gives the correct answer.

The family of Romance languages allows for a direct test of the estimated stabilities against the replacement rates since the proto-language (Latin) is known and the replacement rates can be explicitly computed. The output of the test is threefold: first, we prove that the standard approach which tries to infer the replacement rates trough the estimated stabilities is sound; second, we are able to rewrite the fundamental formula of Glottochronology for a non universal replacement rate (a rate which depends on the item); third, we give indisputable evidence that the stability ranking is far from being the same for different families of languages. This last result is also supported by comparison with the Malagasy family of dialects.

As a side result we also provide some evidence that Vulgar Latin and not Late Classical Latin is at the root of modern Romance languages.

## I. THE PROBLEM

Glottochronology tries to estimate the time at which two languages started to differentiate and, consequently, also the time depth of a family of languages. The assumption is that vocabulary changes at a more or less constant rate so that this time may be inferred from lexical differences.

The idea of measuring language similarity from vocabulary seems to have an ancient root in the work of the French explorer Dumont D'Urville. During his voyages aboard the Astrolabe from 1826 to 1829 he collected word lists of various languages and, in his work about the geographical division of the Pacific [1], he proposed a method for measuring their degree of relation. His approach was to compare pairs of words corresponding to the same item in order to subjectively attribute a number as a measure of their similarity.

The method used by modern Glottochronology, was developed by Morris Swadesh more than sixty years ago [2]. The idea was to measure the similarity of two languages by the proportion of shared cognates in a given Swadesh list of $M$ meanings. The procedure is not so different from D'Urville's approach, the main difference is that the number attributed to a pair of words, which only can be 0 (non cognates) or 1 (cognates), has a precise significance.

Assume that any item (meaning) is labeled by the index $i$ with $i = 1, 2, ..., M$ and any language is labeled by the index $\alpha$ with $\alpha = 1, 2, ..., N$. Also assume that $\alpha_i$ represents the word corresponding to the item $i$ in the language $\alpha$, which implies that $\alpha_i$ indicates a couple of coordinates, i.e., $\alpha_i = (\alpha, i)$. Then, if we define

$$C(\alpha_i, \beta_i) = \begin{cases} 1 & \text{if } \alpha_i \text{ and } \beta_i \text{ are cognate,} \\ 0 & \text{otherwise,} \end{cases} \tag{1}$$

we can compute the overlap $C(\alpha, \beta)$ between two languages $\alpha$ and $\beta$ as

$$C(\alpha, \beta) = \frac{1}{M} \sum_{i=1}^{M} C(\alpha_i, \beta_i), \tag{2}$$

which is, by definition, the proportion of positive cognate matching.

Classical Glottochronology implicitly assumes that the replacement rate $r$ is constant in time and the same for any of the items $i$. Thus, the expected value of any of the random variables $C(\alpha_i, \beta_i)$ is

$$E[C(\alpha_i, \beta_i)] = e^{-rT(\alpha,\beta)}, \tag{3}$$

where $E[\cdot]$ indicates the expected value and where $T(\alpha, \beta)$ is the time distance between languages $\alpha$ and $\beta$. For example, if $\alpha$ and $\beta$ are contemporary languages, then $T(\alpha, \beta)$ is twice the time from their common ancestor, otherwise, if $\alpha$ is the ancestor language of $\beta$ then $T(\alpha, \beta)$ is their time separation.

If the elements in the sum (2) are independent (likely they are, since word replacements usually are independent events) and $M$ is sufficiently large (typically it equals 100 or 200) one can apply the law of large numbers. Therefore, one has that $C(\alpha,\beta) \simeq E[C(\alpha,\beta)]$ which, given (2) and (3), implies

$$C(\alpha,\beta) \simeq \frac{1}{M}\sum_{i=1}^{M} E[C(\alpha_i,\beta_i)] = e^{-rT(\alpha,\beta)}. \tag{4}$$

Reversing this equality one obtains the fundamental formula of Glottochronology:

$$T(\alpha,\beta) \simeq -\frac{1}{r}\ln C(\alpha,\beta). \tag{5}$$

This formula has been derived under the assumption that the value of $r$ is universal, i.e. that it is independent of the item $i$. As well, $r$ must be the same for all pairs of languages in the considered family while, in principle, it could be different for different families.

Its actual value is controversial, nevertheless some proposals have a larger consensus (see, for example [3]). In our opinion it is better to fix the value of $r$ by history. In fact, a single event, which fixes the time distance of a single pair of languages among all pairs, is sufficient to determine $r$. For example, Iceland was colonized by Norwegians about the year 900 CE and therefore the corresponding time distance between Norwegian and Icelandic is about $1.1 \times 2 = 2.2$ millennia.

Amazingly, it has been known for a long time that the rate $r$ is not universal [4–6], but, somehow, this point was neglected by scholars and most of Lexicostatistics studies continued as if the rate were universal. In fact, the fundamental formula (5) only holds if the rate is the same for all items or, at least, its range of variation is very small. We will see that this is not the case, therefore, (5) will be replaced by a generalized version. The estimated value of the time distance $T(\alpha,\beta)$ may change drastically.

Because of these difficulties and inconsistencies, among others, the method of Swadesh was often misunderstood and harshly criticized since the beginning (see, for example, [7]), nevertheless, it is still successfully and widely used in many different contexts [8–12] and it is still subject to improvements and developments [3, 13, 14]. Moreover, significant progress has been made to make the cognate identification process more objective [15–19].

Before presenting our results and methods, we would like to mention that this work is mostly based on an automated version of Swadesh approach which was proposed about ten years ago [20–22]. According to this automated approach, the point that we have modified here is simply the definition (1) which we have replaced by a more objective measure based on Levenshtein distance.

The Levenshtein distance between two words is simply the minimum number of insertions, deletions or substitutions of a single character needed to transform one word into the other. The automated approach [20–22] is based on a normalized Levenshtein distance ($NLD$) which is defined as the Levenshtein distance divided by the number of characters of the longer of the two compared words.

In symbols, given two words $\alpha_i$ and $\beta_i$ corresponding to the same meaning $i$ in two languages $\alpha$ and $\beta$, their normalized Levenshtein distance $D(\alpha_i,\beta_i)$ is

$$D(\alpha_i,\beta_i) = \frac{D_L(\alpha_i,\beta_i)}{L(\alpha_i,\beta_i)}, \tag{6}$$

where $D_L(\alpha_i,\beta_i)$ is the Levenshtein distance between the two words and $L(\alpha_i,\beta_i)$ is the number of characters of the longer of the two. This normalized Levenshtein distance can take any rational value between 0 and 1. Then, in order to compare with the standard Swadesh approach, it is useful do define

$$C(\alpha_i,\beta_i) = 1 - D(\alpha_i,\beta_i), \tag{7}$$

which ideally corresponds to the classical definition of cognacy for which $C(\alpha_i,\beta_i)$ takes the value 1 if the two words are cognate and the value 0 if they are not.

In our case, $C(\alpha_i,\beta_i)$ takes the value 1 only if the two words are identical and it takes the value 0 if they are completely different, but, in general, it takes a rational value between 0 and 1. The definition of the overlap given by (6) and (7) replaces the definition (1), leaving unchanged formulae from (2) to (5) as well all subsequent formulae in this paper.

Surprisingly, while the $C(\alpha_i,\beta_i)$ are definitely different in the two approaches, we recently found out that the corresponding measures of the estimated stability $S(i)$, as well of the overlaps $C(\alpha,\beta)$ are extremely correlated. This will be discussed in a forthcoming paper. Though we use here the $NLD$ approach, all results can be re-obtained by standard approach based on cognacy identification.

The measure of the estimated stability $S(i)$ is based on an idea, which dates back more than fifty years [4–6]. The formulation may vary from an author to another but the main assumption is the same: the more the words corresponding to a given meaning $i$ in different languages are similar, the larger is the stability of this meaning. More recently, this idea was developed and improved [23, 24] and a convincing explanation for the difference of the stability values for different items was also found [25, 26].

The idea was also revisited by using the $NLD$ approach in [27, 29] where it was also shown that the stability ranking is not universal, i.e., it strongly depends on the family of languages under consideration. Ranking, according to [25], is determined by the frequency of word-use and, therefore, it is an interesting tool for the inference of cultural traits of different populations. Moreover, the non-universality of ranking strongly points to the idea that the choice of the items of the Swadesh lists should not be the same for all families but, on the contrary, each family should be studied compiling an ad hoc list.

Let us end this section with a description of our results.

Our concern is that although the idea of inferring the rates of replacement from the estimated stability has been longly and successfully used, nothing ensures that it gives the correct answer. There are different methods for measuring the stability, but the key point is to compare for each meaning $i$ all the words associated to $i$ in the languages of a given family. As already mentioned, the more similar these words are, the higher the stability $S(i)$. In turn, the stability $S(i)$ is assumed to be exponentially proportional to the replacement rate $r_i$ of the words corresponding to the meaning $i$. If this assumption is correct, one can infer the replacement rates (which in general cannot be directly measured) using the stabilities (which can be measured).

The family of Romance languages allows for a direct test of this assumption. In this case, in fact, the estimated stabilities $S(i)$ can be tested against the rates $r_i$ since the proto-language (Vulgar Latin) is known and the the actual rates $r_i$ can be explicitly computed by comparison of modern Romance languages with Vulgar Latin. The conclusion (Section II) is that the idea of estimating the $r_i$ trough the $S(i)$ is sound. Nevertheless, it turns out that the rates of replacement indeed depend on the item $i$. As a side result we also provide some evidence that Late Classical Latin is not at the root of modern Romance languages.

In Section III we compute the distribution of the rates $r_i$ both for those inferred by the estimated stabilities $S(i)$ and those directly computed by comparison with Latin. The result is that both distributions are definitely broad. On the contrary, (5) is derived under the assumption that the replacement rates do not significantly depend on $i$, which implies that the distributions should be narrow. Therefore, it is necessary to rewrite the fundamental formula of Glottochronology (5) in order to incorporate this new result. The new formula, in general, gives a different value for $T(\alpha, \beta)$ as compared to (5).

In section IV we compare the ranking of the replacement rates $r_i$ for different families of languages. First we consider two Romance sub-families and the result is that ranking, even if some correlation is present, is definitively different for the two families. This is a clear evidence that ranking is far from being universal. The result is then confirmed by comparison of the stability ranking for the entire Romance family of languages with the stability ranking for the Malagasy family of dialects. In this case similarity between the two rankings is totally missing.

Our conclusions are contained in section V.

Our database consists of 55 Swadesh lists for contemporary Romance languages (110 items), 2 lists for Late Classical Latin and Vulgar Latin (110 items) and 24 lists for Malagasy dialects (200 items). The main source for Romance and Latin languages was "The Global Lexicostatistical Database, Indo-European family: Romance group" (Version September 2016), which contains annotated Swadesh lists compiled by Mikhail Saenko. For the Malagasy family we have used the database collected by the authors of [11] which consists of Swadesh lists for 24 dialects of Malagasy from all areas of the island. A detailed description including all other sources is contained in the Appendix.

## II. ACTUAL STABILITY, ESTIMATED STABILITY AND REPLACEMENT RATES

According to our database, we consider $N = 56$ languages, 55 of which are Romance languages and one is Latin which can be Late Classical Latin or Vulgar Latin (we tested both but only the second seems to be at the root of the Romance phylogenesis). For any language, the Swadesh lists contain $M = 110$ items.

Any of the $N$ languages is labeled by an integer indicated by a Greek letter (say $\alpha$) with $1 \leq \alpha \leq N$ where $\alpha = 2, ..., N$ correspond to the modern Romance variants and $\alpha = 1$ corresponds to Latin (Late Classical or Vulgar).

Any item (meaning) is labeled by $i$ with $1 \leq i \leq M$. As already mentioned, $\alpha_i$ represents the word corresponding to the item $i$ for the language $\alpha$ which means that $\alpha_i$ indicates a couple of coordinates, i.e., $\alpha_i = (\alpha, i)$.

We may define the actual stability of an item $i$ counting the average overlap between Latin and modern Romance

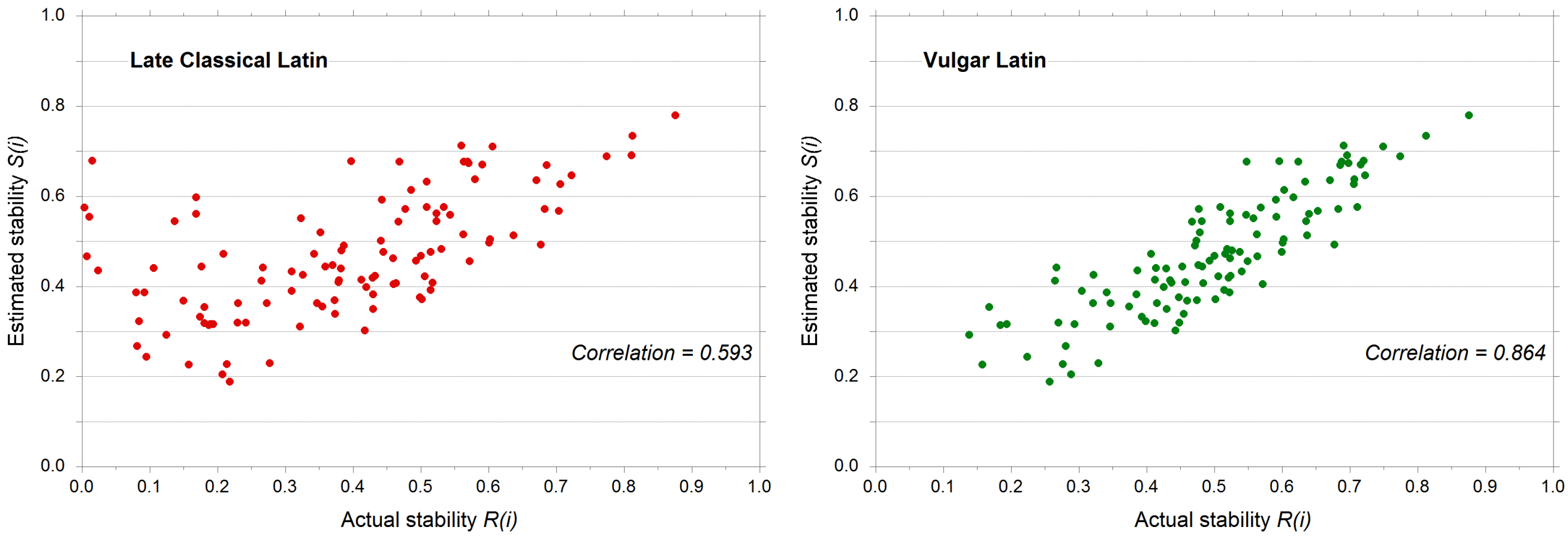

FIG. 1: The estimated stabilities $S(i)$ plotted against the actual stabilities $R(i)$. The estimated stabilities are the same in the two figures. The left figure corresponds to values of the actual stability computed by Late Classical Latin. There is a significant correlation between data, but there is a much stronger correlation if Vulgar Latin is used instead (right figure).

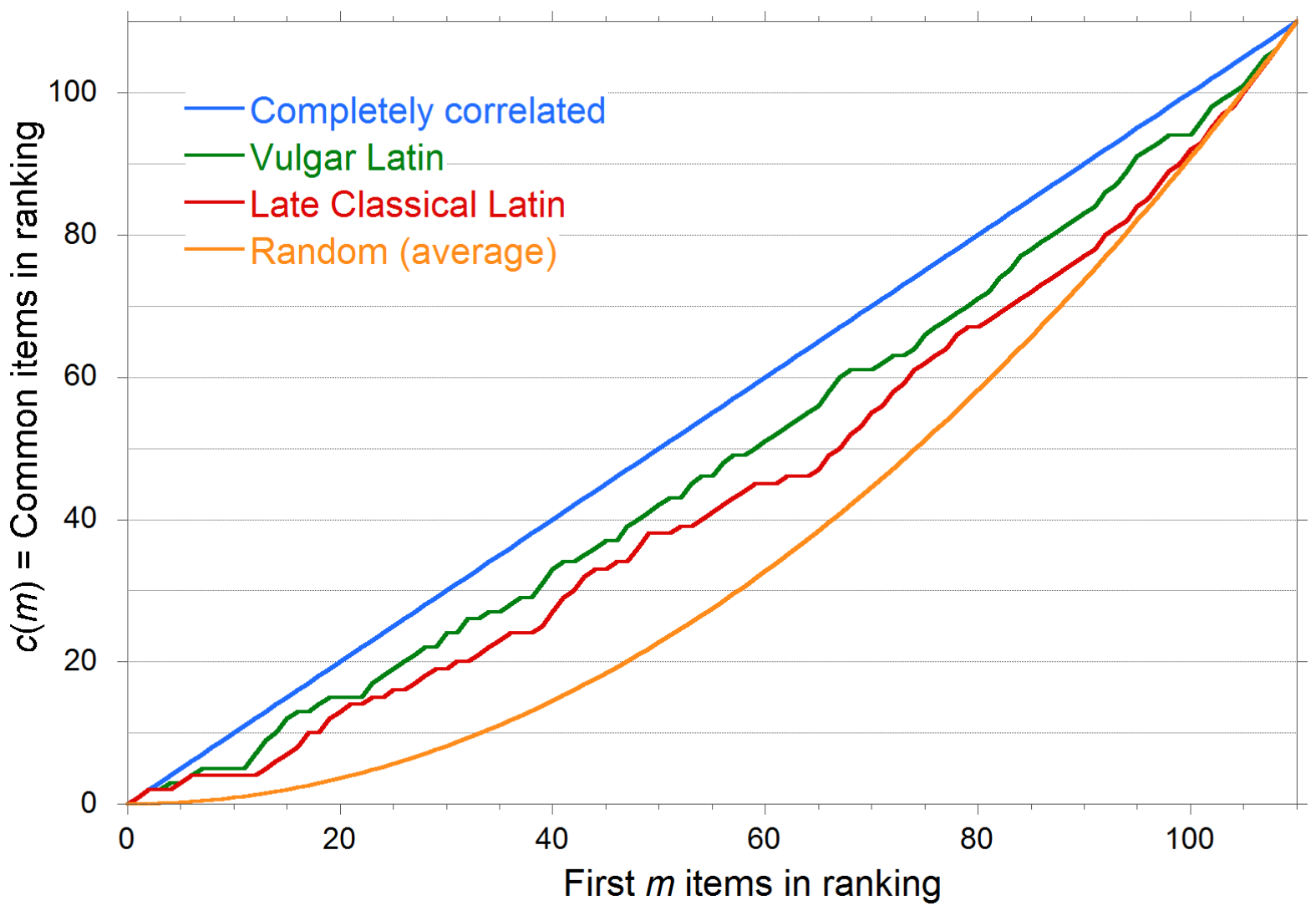

FIG. 2: We considered the first $m \leq M$ items according to the decreasing ranking of the estimated stability $S(i)$ and the first $m$ according to the decreasing ranking of the actual stability $R(i)$. These two shorter lists have $c(m) \leq m$ items in common. If the two lists were independently compiled choosing randomly $m$ items among $M$ they would have in average a number $c(m) = \frac{m^2}{M}$ of common items (gold line). On the contrary, if the ranking of the $M$ estimated stabilities and of the $M$ actual stabilities were exactly the same, they would obviously have a number $c(m) = m$ of common items (blue line). The reality is in the middle. Vulgar Latin (green line) has a $c(m)$ which is much closer to the straight line $c(m) = m$ than Late Classical Latin (red line).

languages as

$$R(i) = \frac{1}{N-1} \sum_{\beta=2}^{N} C(1_i, \beta_i) \tag{8}$$

where the sum goes on all modern Romance languages ($\beta = 2, .....N$).

Notice that with this definition all the $R(i)$ are in the interval $[0, 1]$. Also, notice that only pairs of words corresponding to the same item $i$ are used in this definition and, therefore, the rate of replacement $r_i$, corresponding to item $i$, is the same for all words. According to the probabilistic digression in previous section one has that all the $C(1_i, \beta_i)$ are equally distributed with average

$$E[C(1_i, \beta_i)] = e^{-r_i T}, \tag{9}$$

where $T = T(1,\beta)$ is the time distance between each modern Romance languages $\beta$ and Latin.

This is the contrary of the situation usually faced by Glottochronology since in our case the time $T(1,\beta)$ is known while the rate of replacement is not. In fact, for all Romance languages in the above formula, $T(1,\beta)$ equals $T$ which is the time distance between the ancient Rome period (Latin was spoken) and the present (Romance languages are spoken). More precisely, if Late Classical Latin is considered one has that $T$ is about 1.85 millennia, while if Vulgar Latin is considered one has that $T$ is about 1.5 millennia. This is because the authors of "The Global Lexicostatistical Database" based their Late Classical Latin on the works of Apuleius (around 150 CE) integrated in difficult cases by "Noctes atticae" by Aulo Gellio (159-170 CE) and "Satyricon" by Petronius (1st century CE), so that $T = 1850$ years is explained. Vulgar Latin, on the contrary, is the language which was spoken by the *vulgus* at the end of the Roman empire so that the value $T = 1500$ years is justified in this case.

We tested both versions of Latin and we found, as it will be clear in the following, that Vulgar Latin can be easier placed at the root of Modern Romance Languages.

This time, in order to find out our result from formulae (8) and (9) by means of the law of large numbers, we have to assume that $N$ is sufficiently large. We obtain

$$R(i) \simeq \frac{1}{N-1}\sum_{\beta=2}^{N} E[C(1_i,\beta_i)] = e^{-r_i T} \quad \rightarrow \quad r_i \simeq -\frac{1}{T}\ln R(i). \tag{10}$$

Since the value of $T$ is given, we found out that computing the $M = 110$ actual stabilities $R(i)$ by data using (8) allows to obtain the corresponding replacement rates $r_i$.

The big problem is that in most cases, differently from the Romance languages case, only contemporary languages are available, while the ancestor language (proto-language) is unknown. In this case, not only one cannot compute the overlaps $C(1_i,\beta_i)$ but also the time distance $T$ from the proto-language is unknown. As a result the replacement rates $r_i$ cannot be computed by the method explained above.

It is clear that in this more common case it is necessary to infer the rates $r_i$ by some procedure that only involves living languages.

We show now that the estimated stability is a good tool to obtain accurate approximations for the rates $r_i$. Since the estimated stability $S(i)$ must be computed using only contemporary languages, the simplest choice is to define it as

$$S(i) = \frac{2}{(N-1)(N-2)}\sum_{\alpha>\beta\geq 2} C(\alpha_i,\beta_i) \tag{11}$$

where the sum goes on all $(N-1)(N-2)/2$ possible language pairs $\alpha$, $\beta$ between the $N-1$ Romance languages.

The estimated stability $S(i)$ measures the average overlap (for the meaning $i$) between all pairs of contemporary Romance languages. This quantity is candidate for replacing the actual stability $R(i)$ (which measures the average overlap between Latin and all modern Romance languages) in those cases in which the proto-language is unknown.

Reasonably $S(i)$ is smaller for those meanings $i$ with a higher rate of lexical evolution since they tend to remain more similar in time, but our goal is more ambitious, i.e., we would like that the estimated stability and the actual stability were strongly correlated. Only in this case $S(i)$ could be used in place of $R(i)$ for estimating the replacement rate $r_i$.

Therefore, all we have to do is to test the correlation between the estimated stability and the actual stability. We plotted $S(i)$ against $R(i)$ (Fig. 1) using Late Classical Latin (left) and Vulgar Latin (right). If the data in the figure were perfectly aligned (one is a linear transformation of the other) we could conclude that estimated stability (11) and actual stability (10) were equivalent. This is not the case, nevertheless, there is a strong correlation between the $S(i)$ and $R(i)$ variables. This means that we can use the $S(i)$ in order to estimate the $r_i$.

It is important to observe that the correlation is much stronger for Vulgar Latin with respect to Late Classical Latin. This probably means that Vulgar Latin is at the root of the phylogenesis of Romance languages while Late Classical Latin is a side branch. This view is coherent with the common belief that the Late Classical Latin was an artificial literary continuation of the Archaic Latin without a precise correspondence with the evolution of the spoken language. Indeed, it is widely underlined in literature that spoken language and the written language already started to diverge about a century before CE.

Before proceeding with our extrapolation of the $r_i$ from the estimated stabilities, let us confirm the result obtained by measuring the correlations with another measure. Given the list with $M = 110$ items $i$, we order them in a decreasing ranking according to $R(i)$ and we do the same for the $S(i)$. If we consider the first $m \leq M$ items according to the ranking $S(i)$ and the first $m$ according to ranking $R(i)$, we can count the number $c(m) \leq m$ of common elements in these two shorter lists.

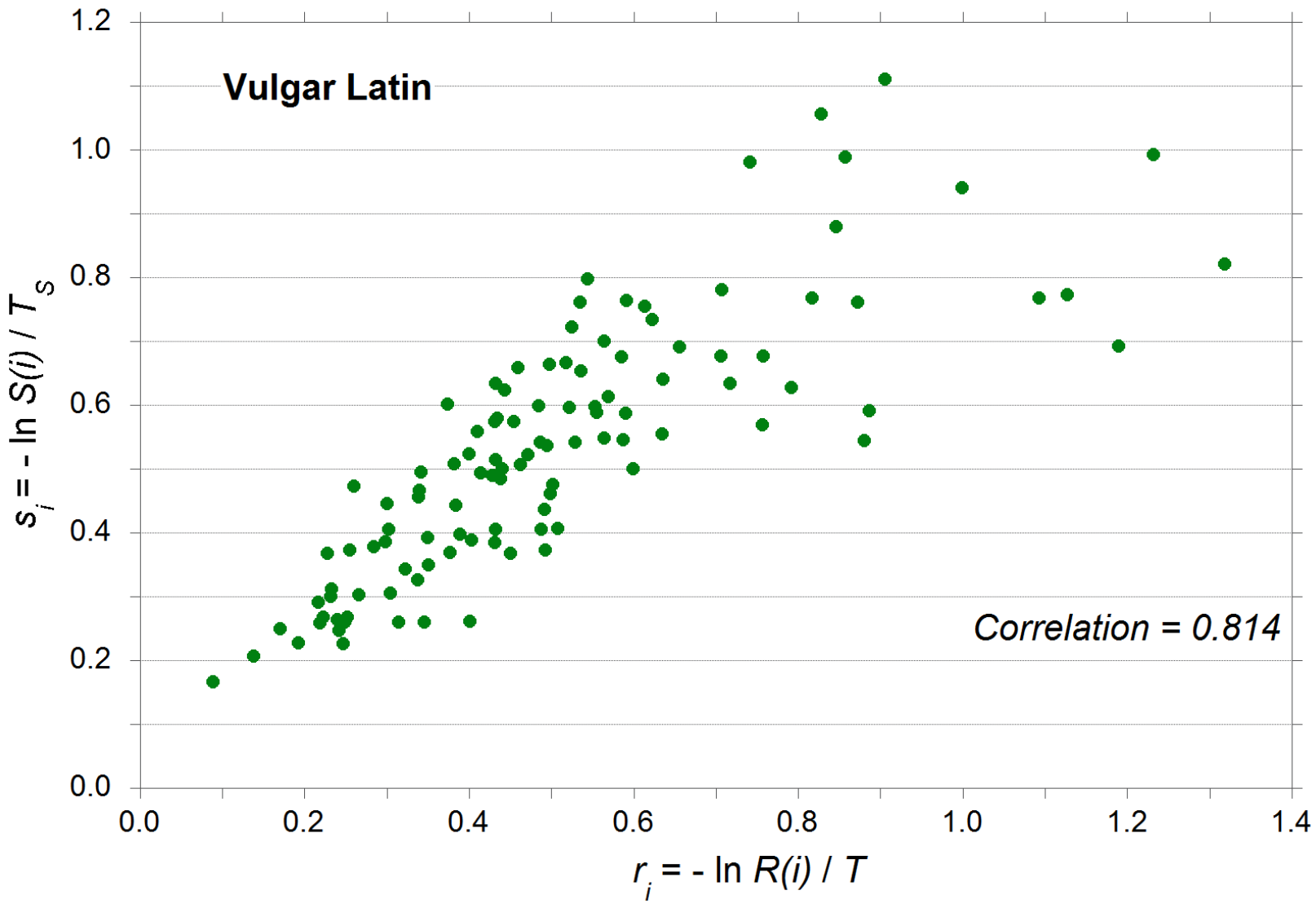

FIG. 3: The $s_i = -\frac{1}{T_s}\ln S(i)$ are plotted against the $r_i = -\frac{1}{T}\ln R(i)$. The $R(i)$ are computed using Vulgar Latin so that $T$ is fixed by history to be about 1.5 millennia. The value of $T_s$ is also arbitrarily fixed to be equal to 1.5. There is a strong correlation and the data are roughly aligned.

If the two lists were independently compiled choosing $m$ items randomly among $M$ they would have in average a number $c(m) = \frac{m^2}{M}$ of common items. On the contrary, if the ranking of the $M$ estimated stabilities and of the $M$ actual stabilities were exactly the same, they would obviously have a number $c(m) = m$ of common items for any $m$. We plotted the result in Fig. 2, which shows that reality is in the middle. Again, Vulgar Latin gives a better performance compared to Late Classical Latin, in fact, the value of $c(m)$ computed with Vulgar is systematically larger than the $c(m)$ computed with Late Classical Latin.

Our conclusion is that the actual stabilities $R(i)$ computed from Vulgar Latin are very correlated to the estimated stabilities $S(i)$. Therefore, in all cases in which we don't know the proto-language (the $R(i)$ cannot be computed) we can try to extrapolate the $r_i$ from the observables $S(i)$. But how?

The $r_i$ are computed using the equality at the right in formula (10). On the other hand, we have shown that the $S(i)$ defined in (11) and the $R(i)$ defined in (8) are highly correlated. Therefore, in analogy with (10), we are tempted to define the variables $s_i$ as

$$s_i = -\frac{1}{T_s}\ln S(i), \qquad (12)$$

with the hope that they are a good estimate of the $r_i$ (which, in general, cannot be directly measured).

The parameter $T_s$ is arbitrary at this stage since the time depths of the family is supposed to be unknown (unless the proto-language is historically attested as for the Romance family). Fortunately, the mathematical correlation between the $s(i)$ and the $r(i)$ does not depend on $T_s$ so that we can make any arbitrary preliminary choice for this parameter. The strong correlation between the $r_i$ and the $s_i$ is shown by Fig. 3 (where we used $T_s = T = 1.5$ millennia). Moreover, this figure also allows us to roughly assume

$$r_i \simeq \lambda s_i. \qquad (13)$$

The parameter $\lambda$ remains unfixed at this stage since the correct value of $T_s$ is unknown. On the other side we cannot fix it by a plot as in Fig. 3 since the proto-language is unknown and the $r_i$ cannot be computed. Nevertheless, as we will show in next section, there is a way to fix $\lambda$ using only contemporary languages.

Let us briefly summarize the results of this section:

- if the proto-language is known the replacement rates $r_i$ can be computed by the equation at the right side of (10) where the $R(i)$ are given by (8);
- in most cases the proto-language is unknown, but one can anyway compute the $s_i$ using formulae (11) and (12). The time parameter $T_s$ in (12) is arbitrary;

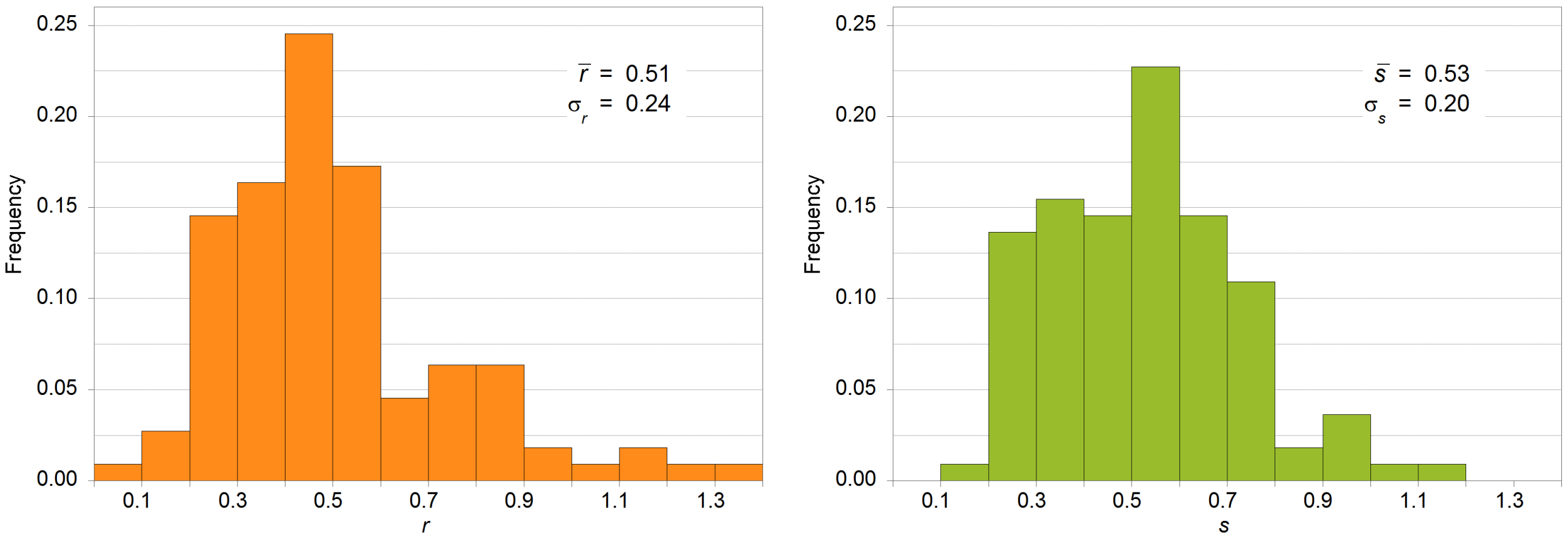

FIG. 4: The distribution of the $r_i$ (left) and of the $s_i$ (right). The frequency is the number of occurrences in an given interval of size 0.1 divided by the total number $M = 110$ of occurrences. The interesting fact is that both distributions remain large between 0.2 and 0.8 meaning that rates of different items may typically differ by a factor two or more.

- we have shown that $r_i \simeq \lambda s_i$. The value of the parameter $\lambda$ remains unfixed at this stage as a consequence of the arbitrariness of $T_s$. We will show in next section how to fix its value using only contemporary languages.

## III. DISTRIBUTION OF RATES AND TIME DISTANCES

It is already clear from Fig. 3 that the replacement rates $r_i$, deduced comparing Latin with Romance languages (as well the replacenment rates $s_i$, deduced by comparing only Romance Languages) are very different for different items $i$.

This fact is even more evident in Fig. 4 were the distributions of the rates $r_i$ and $s_i$ are plotted. In fact, both distributions remain large in the interval between 0.2 and 0.8 meaning that rates of different items may typically differ by a factor two or more. This is really astonishing given that the fundamental formula of Glottochronology is derived with the assumption that the rate of replacement is universal i.e., that all $r_i$ are equal.

At this point the question is: how should we modify the fundamental formula of Glottochronology in order to accommodate for this fact? What one can do, is to assume again that the overlap may be replaced by its expected value so that

$$C(\alpha,\beta) \simeq \frac{1}{M}\sum_{i=1}^{M} E[C(\alpha_i,\beta_i)] = \frac{1}{M}\sum_{i=1}^{M} e^{-r_i T(\alpha,\beta)}, \tag{14}$$

where, unlike formula (4), the rates depend on the item $i$.

If the actual rates $r_i$ can be preliminarily computed from data (the proto-language is known), the time distances $T(\alpha,\beta)$ can be obtained by inverting the above formula. For each pair of languages one gets $T(\alpha,\beta)$ as a function of $C(\alpha,\beta)$ (which in turn is computed from data). The inversion is always possible since the relation between $C(\alpha,\beta)$ and $T(\alpha,\beta)$ is clearly monotonic.

In case the actual rates cannot be computed (the proto-language is unknown) one can use the estimated rates $s_i$ in place of the $r_i$. Assuming proportionality between the two set of variables. i.e., $r_i = \lambda\, s_i$, the above formula rewrites as

$$C(\alpha,\beta) \simeq \frac{1}{M}\sum_{i=1}^{M} E[C(\alpha_i,\beta_i)] = \frac{1}{M}\sum_{i=1}^{M} e^{-\lambda s_i T(\alpha,\beta)}. \tag{15}$$

Given that all the $s_i$ can be preliminarily computed (even in absence of a known proto-language), also this formula can be inverted and the time distances $T(\alpha,\beta)$ between two languages can be still obtained as a function of $C(\alpha,\beta)$. All the $T(\alpha,\beta)$ also depend on the single parameter $\lambda$ which is not fixed at this stage, but it is sufficient to know the time distance of a single pair of languages to determine it. This fixing has to be specific for the family which is considered, for example for the Romance family one knows that Sardinian and Sicilian started to diverge at the fall of the Roman empire, about the year 500 CE and, therefore, the corresponding time distance between the two languages is about $1.5 \times 2 = 3$ millennia. In this work, we don't need to compute the divergence times $T(\alpha\beta)$ for Romance languages, the problem is studied in detail in a forthcoming article [28]. In case the fixing is not possible because there are not historically attested dates for the beginning of a divergence, the global scale of time remains arbitrary but all relative time distances are still computable.

We can summarize the results in this section as follows;

- the inversion of formula (15), which, in general, has to be done numerically, gives the time distance $T(\alpha, \beta)$ in terms of $C(\alpha, \beta)$ and of the $s_i$, which are all quantities computable from data even in absence of a known proto-language. Therefore, the inverted formula is a generalization of the fundamental formula of Glottochronology (5);
- all the $T(\alpha, \beta)$ also depend on $\lambda$ which is not fixed at his stage, but it is sufficient to know the time distance of a single pair of languages to fix it. In case this is not possible, the global scale of time remains arbitrary;
- the time distances $T(\alpha, \beta)$ may differ consistently from those computed with the standard fundamental formula which can be eventually re-obtained replacing all the $s_i$ in (15) by the average $\bar{s} = \frac{1}{M} \sum_{i=1}^{M} s_i$. This is reasonable only if the $s_i$ have a narrow distribution, which seems not to be the case.

In order to have a qualitative perception on how the fundamental formula may change drastically, assume, for example, that the distribution of the $s_i$ can be approximated by a probability density

$$\rho(s) = \frac{s^{Z-1}}{\Gamma(Z)\, P^Z} e^{-\frac{s}{P}}, \tag{16}$$

where $\Gamma$ is the standard Gamma function and the positive parameters $Z$ and $P$ are chosen such that the average $ZP$ and the deviation $\sqrt{Z}\,P$ coincide with average $\bar{s} = 0.53$ and deviation $\sigma_s = 0.20$ of the distribution in the right side of Fig. 4. Approximately, $Z = 7.0$ and $P = 0.076$.

We have chosen this probability density because the associated frequency has a shape which is similar to the one in the right side of Fig. 4, but any other density with similar characteristics could equally be used. In fact, the purpose is simply to give a qualitative understanding of the result of the inversion of formula (15) which can be easily and exactly performed by numerical tools without the need of introducing distributions. One gets:

$$C(\alpha, \beta) \simeq \int_0^\infty \rho(s) e^{-s\lambda T(\alpha,\beta)} ds = \frac{1}{[1 + \lambda P\, T(\alpha, \beta)]^Z} \quad \rightarrow \quad T(\alpha, \beta) \simeq \frac{1}{\lambda P} \left( \frac{1}{[C(\alpha, \beta)]^{\frac{1}{Z}}} - 1 \right) \tag{17}$$

where the first equality at the left side is the continuous approximation of (15), while the second equality is obtained by computing the integral with $\rho(s)$ given by (16).

The right side of the above equation, which is computed by inverting the equality at the right side, has a definitely different shape as compared to (5). For example the first function diverges logarithmically for small values of $x$ while the second has a power law divergence.

## IV. NON UNIVERSALITY OF THE STABILITY RANKING

Stability ranking is not universal, but it strongly depends on the family of languages under consideration. This fact was already evidenced in [27, 29]. We provide here further evidence on a more quantitative basis.

In order to compare the rankings, we compiled the lists with the first $m \leq M$ items according to the decreasing ranking of the estimated stability $S(i)$ for different families and sub-families. Then, we measured the number $c(m) \leq m$ of common items of two lists. In both cases we considered, we found a result closer to $c(m) = \frac{m^2}{M}$ than $c(m) = m$. The value $c(m) = m$ corresponds to the case in which rankings are identical while $c(m) = \frac{m^2}{M}$ is the average value of the number of common items of two lists which are independently composed choosing randomly $m$ items among $M$.

This result is depicted in Fig. 5 where the Western Romance sub-family was compared with the Eastern Romance sub-family and the Romance family with the Malagasy family of dialects.

The 14 Eastern and the 41 Western Romance languages are separated by the La Spezia-Rimini line. More precisely the eastern sub-family contains the Italian variants below La Spezia-Rimini line and those variants spoken in eastern Europe, while the Western sub-family contains all the others.

The Romance family contains all the 55 languages of the dataset, while for the Malagasy family we have used all the 24 dialects of Malagasy collected by the authors of [11] (see the Appendix). Since the comparison was made between Malagasy and Romance families, we used only the 103 common items which are both in the 200 item lists of Malagasy dialects and in the 110 items lists of Romance languages.

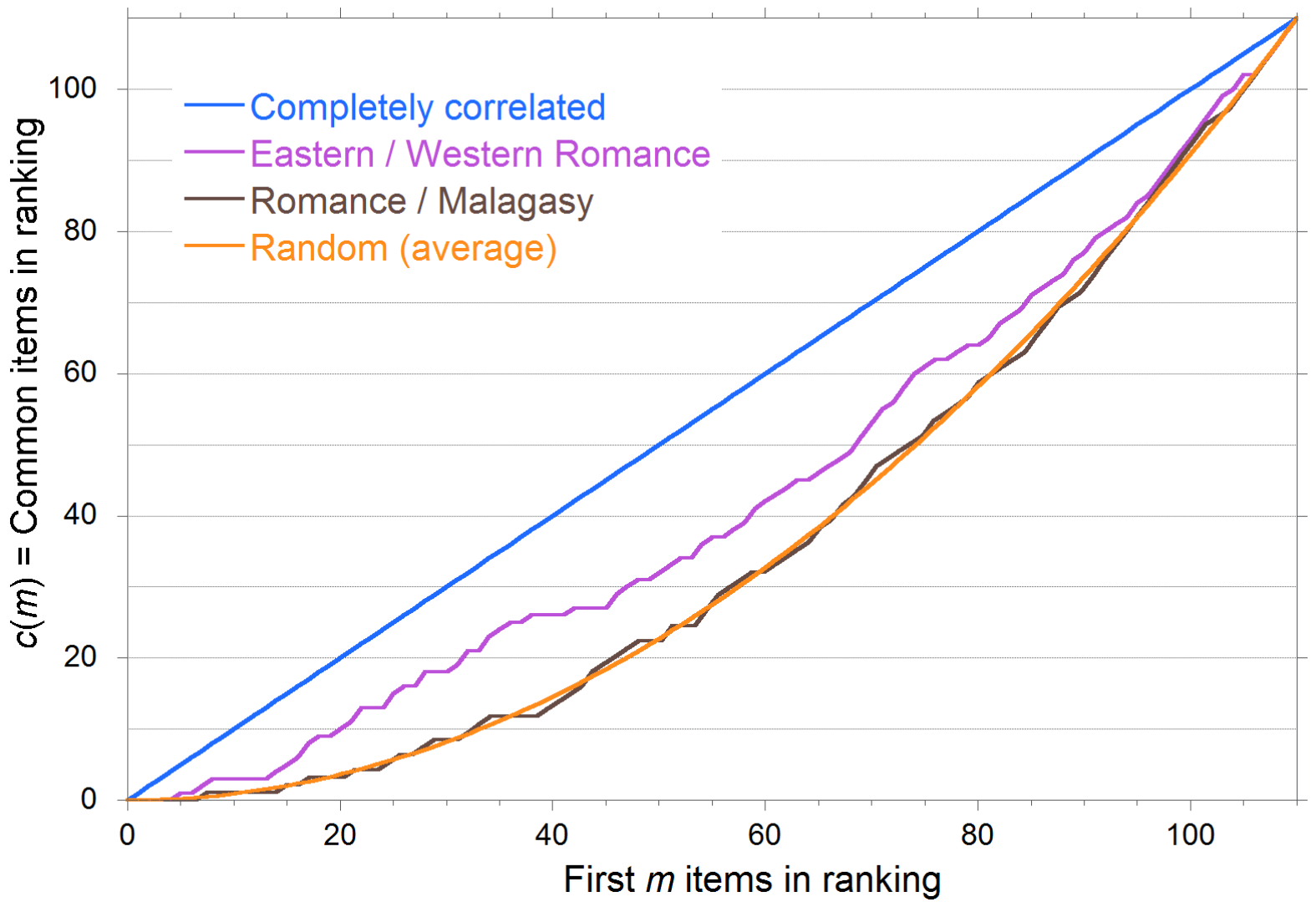

FIG. 5: We considered the lists with the first $m \leq M$ items according to the decreasing ranking of the estimated stability $S(i)$ for the Western Romance sub-family, the Eastern Romance sub-family, the Romance family and the Malagasy family. The number $c(m)$ of common items of two lists is plotted for Eastern/Western (purple line) and Romance/Malagasy (brown line). The Malagasy/Romance $c(m)$ is extremely close to $c(m) = \frac{m^2}{M}$ (gold line), while the Easter/Western $c(m)$ is intermediate between $c(m) = \frac{m^2}{M}$ and $c(m) = m$ (blue line), but still closer to the first. This means that Romance and Malagasy rankings have a resemblance which is entirely imputable to chance. On the contrary, there is a correlation, although not very large, between Western and Eastern Romance rankings.

The Malagasy/Romance $c(m)$ is extremely close to the average $c(m) = \frac{m^2}{M}$ of completely uncorrelated rankings, while the Easter/Western $c(m)$ is intermediate between $c(m) = \frac{m^2}{M}$ and completely correlated $c(m) = m$, but still closer to the first.

Therefore, comparison very clearly shows that ranking is completely different for Malagasy and Romance families and at least partially different for Eastern and Western Romance sub-families.

Ranking, according to [25], is determined by the frequency of word-use which, in turn, may depend on the specific cultural traits of a given population. Therefore, it is not astonishing that the coincidence of rankings is higher between Western and Eastern Romance sub-families as compared to the coincidence of rankings between Romance and Malagasy families. In the first case the reference cultural traits are likely more similar than in the second.

The clear conclusion of this simple analysis is that ranking is significantly different for different families since in both considered cases the real $c(m)$ is definitely not equal and not even close to the value $c(m) = m$ which corresponds to equal ranking.

This fact that ranking is not universal should be taken into account when one has to make a choice for the items to be included in a Swadesh list. In past (and present) research, on the contrary, the Swadesh list were identical for different linguistic families. This means that the specific cultural traits are not considered relevant variables for the choice of the items to be included in the list. We think, on the contrary, that it would be reasonable to include those items which are ranked first according to stability computed for a specific family. In this way, the items included in a Swadesh list would be different for different families and the choice would privilege in all cases those which are most stable.

This proposal also has practical side, consider the case in which one is dealing with a family of Malagasy dialects, in this case it is of little help to consider the words corresponding to the items "snow" or "ice" which appear in the standard Swadesh list, it would be more reasonable to include instead "rice" or "pirogue". In general, the best one can do is to consider a large vocabulary and choose those (100 or 200) words wich are ranked first according to their stability specifically computed for that family.

## V. CONCLUSIONS

In this paper we have proven that the word replacement rates associated to the meanings $i$ are not all equal but they may differ by a factor two or more. These replacement rates can be directly computed only if the proto-language of a family is known as for example for the Romance family. In this care the replacement rates $r_i$ are obtained by vomparison of Latin with contemporary Romance Languages.

We have shown that in case the proto-language is unknown, the replacement rates can be reasonably inferred using only the vocabulary of the living languages of a family (the $s_i$).

This experimental result was possible since we considered the Romance family which has the fortunate characteristics to have a known proto-language (Vulgar Latin). Therefore, we could make an experimental test comparing the actual stabilities $R(i)$, which could be explicitly computed comparing modern Romance languages with Vulgar Latin, with the estimated stabilities $S(i)$ which are computed comparing only Romance languages. The result was a strong correlation between estimated and actual stabilities.

Since the actual stabilities $R(i)$ are simply related to the rates $r_i$ and the estimated stabilities $S(i)$ are simply related to the rates $s_i$, the conclusion is that the idea of estimating the replacement rates by using the $S(i)$ in place of the $R(i)$ is sound. This result is very useful since the estimated stabilities $S(i)$ can be computed even when the proto-language is unknown, in fact, they are obtained comparing only living languages (Romance languages in our case).

We have also computed the distribution of the replacement rates for the Romance family, both for those inferred by the $S(i)$ (the $s_i$) and for the ones computed from the actual stabilities $R(i)$ (the $r_i$) . The result is that both distributions are definitely broad, while the fundamental formula of Glottochronology implicitly assumes that all the replacement rates are equal. This fact led us to rewrite the fundamental formula of Glottochronology for the case in which the rates $r_i$ strongly depend on the item $i$. In our opinion, this is a necessary step for future improvements of the phylogenetic analysis of all family of languages.

The new formula not only requires to compute the overlap $C(\alpha, \beta)$ for a pair of languages, but also all the estimated rates $s_i$. The weak point is that in order to compute the $s_i$ it is necessary to have a sufficiently large number of languages of a family in order that the law of large numbers can be applied.

In the future, comparing the empirical distribution of the estimated rates for different families, it will probably be possible to infer a universal theoretical distribution. In this case, (15) would be substituted by an equation with the form of the first equality at the left of (17) with $\rho(s)$ representing the inferred universal distribution. In this way the inversion would not require the computation of all the single $s_i$ and it would lead to a new and more precise version of the fundamental formula of Glottochronology.

In the last section, by a simple statistical tool, we gave indisputable evidence that the stability ranking is definitively not the same for different families of languages. This last result was also supported by comparison with the Malagasy family of dialects. The non-universality of the ranking strongly points to the idea that the choice of the items of the Swadesh lists should not be the same for all families but, on the contrary, each family should be studied compiling an ad hoc list according to its specific ranking.

### Acknowledgements

Michele Pasquini was partially supported by the grant "Linguistica Quantitativa", Prot. n. 1755, 14th June 2017, Repertorio n. 98/2017 from the Università degli Studi dell'Aquila.

## APPENDIX

Our database consists of 55 Swadesh lists for contemporary Romance languages (110 items), 2 lists for Late Classical Latin and Vulgar Latin (110 items) and 24 lists for Malagasy dialects (200 items).

Our main source for Romance and Latin languages was "The Global Lexicostatistical Database, Indo-European family: Romance group" (Version September 2016), which contains annotated Swadesh lists compiled by Mikhail Saenko, ("GLD-Romance" in the following). The GLD-Romance is part of the "The Global Lexicostatistical Database" which can be consulted at http://starling.rinet.ru/new100/main.htm .

GLD-Romance lists contain all of the elements of the Swadesh original 100 items list, plus 10 items of the Swadesh 200 item list which were added for special correctional purposes. In order to apply automatic algorithms for data analysis we treated GLD-Romance according to the following operations:

1) if an entry admits more than one synonym, or more than one interchangeable variant, we accepted only the most common among the alternatives with a specific transcription (see point 3);

2) GLD-Romance includes 58 languages. We accepted 53 of them, neglecting "Vercellese Piemontese", "Rapallo Ligurian" (lack of a specific transcription, see point 3), "Archaic Latin", "Old Italian", "Old French" (unnecessary for our research);

3) the entries are transliterated into a unified transcriptional system (a GLD variant of standard IPA), but also report a specific transcription for the given language. We accept this second transcription, with some simplifications:

   - the following substitutions were performed: ā → a, æ → ae, {ç or č} → c, ē → e, đ → d, ɣ → g, ł→ l, ŋ → n, {ɔ or ō or ö} → o, œ → oe, {š or ș} → s, {ū or ü} → u, ž → z,

   - all other characters (accents, graphic or phonetic marks, etc.) different from a white space and not included in modern English alphabet of 26 letters, were omitted;

4) some minor changes were applied to specific entries:

   - the entries of the verb "to walk (to go)", which were recorded by GLD-Romance in the third-person singular present tense form, were transposed to the infinitive form,

   - all the Romanian verbs were rewritten without the initial "a " e.g., "a arde" (to burn) was rewritten as "arde",

   - three Italian entries ("belly", "lie", "snake") were changed in order to be more close to current spoken language;

5) all missing entries were recovered from different sources:

   - Dalmatian: "The Info List - Dalmatian Language", http://www.theinfolist.com/php/SummaryGet.php?FindGo=dalmatian_language ,

   - Megleno Romanian: "Glosbe", https://it.glosbe.com/it/ruq/meglenorumeno ,

   - Ravennate Romagnol: "Vocabolario Romagnolo-Italiano by Libero Ercolani", https://archive.org/details/ vocabolarioromag00ercouoft and "Vocabolario Romagnolo-Italiano by Antonio Morri", https://books.google.it/books?id=e81FAAAAcAAJ ,

   - Provençal Occitan: "IELex - Indo-European Lexical Cognacy Database", http://ielex.mpi.nl./language/Provencal/ ,

   - Savoyard Franco-Provencal: "Dikchonéro Fransé-Savoyâ - Dictionnaire Français-Savoyard by Roger Viret", https://www.arpitania.eu/aca/documents/Dictionnaire_Viret_Francais_Savoyard.pdf ,

   - four missing specific transcription of "claw (nail)" for "Rumantsch Grischun", "Sursilvan Romansh", "Surmiran Romansh" and "Vallader Romansh", were deduced by analogy from GLD-IPA transliteration,

   - three missing "Late Classical Latin" entries ("bark", "fat n." and "red") were chosen after a careful reading of GLD-Romance notes (respectively: "cortex", "adips" and "ruber").

Our database was completed by three modern Romance languages ("Lenguadocian Occitan", "Guascon Occitan" and "Aragonese") obtained from: https://en.wiktionary.org/wiki/Appendix:Swadesh_lists#Indo-European_languages .

We also added a "Vulgar Latin" in the accusative declination, mainly based upon the following sources: "Appendix Probi" (V-VI century); the book "Vulgar Latin" by Jozsef Herman, translated by Roger Wright, which was first published in France as "Le latin vulgaire", Paris, 1967; "Comparative Grammar of the Romance Languages - A Handbook for Exploring Vulgar Latin & the Romance Languages", http://www.nativlang.com/romance-languages/romance-dictionary.php ; "The Reichenau Glosses", compiled in the VIII century in Picardy; web sources as http://it.wikipedia.org/wiki/Latino_volgare and http://en.wikipedia.org/wiki/Vulgar_Latin_vocabulary . A few words were also derived by a paper of S. Starostin [3], by some fragments of texts of Caelius Aurelianus (V century) and by the notes in http://starling.rinet.ru/cgi-bin/response.cgi?root=new100&morpho=0&tableMode=tables&limit=-1&basename= new100\ier\rom+new100\ier\rom&compare=plt+apl .

In conclusion, our database on Romance languages contains 57 complete lists of 100 items, including 55 modern Romance languages and 2 variants of Latin. These languages are: Megleno Romanian, Istro Romanian, Aromanian, Romanian, Dalmatian, Friulian, Gardenese Ladin, Fassano Ladin, Rumantsch Grischun, Sursilvan Romansh, Surmiran Romansh, Vallader Romansh, Lanzo Torinese Piemontese, Barbania Piemontese, Carmagnola Piemontese, Turinese Piemontese, Bergamo Lombard, Plesio Lombard, Ravennate Romagnol, Ferrarese Emiliano, Carpigiano Emiliano, Reggiano Emiliano, Genoese Ligurian, Stella Ligurian, Venice Venetian, Primiero Venetian, Bellunese Venetian,

Standard Italian, Grosseto Italian, Foligno Italian, Neapolitan, Logudorese Sardinian, Campidanese sardinian, Palermitan Sicilian, Messinese Sicilian, Catanian Sicilian, South-Eastern Sicilian, Central Catalan, North-Western Catalan, Minorcan Catalan, Castello de la Plana Catalan, Valencia Catalan, Manises Catalan, Castilian Spanish, Asturian, Standard Portuguese, Galician, Aragonese, Lenguadocian Occitan, Guascon Occitan, Provencal Occitan, Savoyard Franco-Provencal, Standard French, Picard, Walloon, Late Classical Latin, Vulgar Latin.

The entire database can be dowloaded at http://people.disim.univaq.it/~serva/languages/55+2.romance.htm .

For compiling the Swadesh list of Late Classical Latin, the GLD authors mainly used the texts of Apuleio (about 150 CE) and, at a lesser extent, the "Noctes Atticae" by Aulo Gellio (159-170 CE). For this reason we fix the Late Classical Latin around 150 CE. According to the main sources that we used for compiling the Swadesh list of Vulgar Latin we approximatively fix the Vulgar Latin around 500 CE.

For the Malagasy family we have used the database collected by the authors of [11] which consists of 200-item Swadesh lists for 24 dialects of Malagasy from all areas of the island. These dialects are (name of the variant and corresponding town): Antambohoaka (Mananjary), Antaisaka (Vangaindrano), Antaimoro (Manakara), Zafisoro (Farafangana), Bara (Betroka), Betsileo (Fianarantsoa), Vezo (Toliara), Sihanaka (Ambatondranzaka), Tsimihety (Mandritsara), Mahafaly (Ampanihy), Merina (Antananarivo), Sakalava (Morondava), Betsimisaraka (Fenoarivo-Est), Antanosy (Tolagnaro), Antandroy (Ambovombe), Antankarana (Vohemar), Masikoro (Miary), Antankarana (Antalaha), Sakalava (Ambanja), Sakalava (Majunga), Sakalava (Maintirano), Betsimisaraka (Mahanoro), Antankarana (Ambilobe), Mikea (Analamena).

This database is available at http://people.disim.univaq.it/~serva/languages/malgasce.htm .

---